\documentclass[12pt,letterpaper]{article}
\usepackage[a4paper, total={7in, 10in}]{geometry}

\usepackage{graphicx}
\usepackage{helvet}
\usepackage{authblk}
\usepackage{hyperref}
\usepackage{amsmath} 
\usepackage{amssymb} 
\usepackage{orcidlink} 
\usepackage[super,comma,sort&compress]  
   {natbib}
\usepackage[right]{lineno} \linenumbers
\usepackage{graphicx}

\usepackage{booktabs}
\usepackage{multirow}
\usepackage{siunitx}
\usepackage{algorithm}
\usepackage{algorithmic}
\usepackage{url}

\makeatletter
\renewcommand{\maketitle}{\bgroup\setlength{\parindent}{0pt}
\begin{flushleft}
  \textbf{\@title}
  
  \@author
\end{flushleft}\egroup}
\makeatother

\title{EfficientMIL: Efficient Linear-Complexity MIL Method for WSI Classification}
\date{}

\author[1,2,5]{Chengying She}
\author[4,5]{Chengwei Chen}
\author[3]{Dongjie Fan}
\author[1,6,*]{Lizhuang Liu}
\author[4,*]{Chengwei Shao}
\author[4,*]{Yun Bian}
\author[1,2]{Ben Wang}
\author[1,2]{Xinran Zhang}

\affil[1]{Shanghai Advanced Research Institute, Chinese Academy of Sciences, Shanghai, China}
\affil[2]{University of Chinese Academy of Sciences, Beijing, China}
\affil[3]{North University of China, Taiyuan, China}
\affil[4]{Department of Radiology, Changhai Hospital, Shanghai, China}
\affil[5]{These authors contributed equally}
\affil[6]{Lead contact}

\affil[*]{Correspondence: liulz@sari.ac.cn (L.L.), chengweishaoch@163.com (C.S.), bianyun2012@foxmail.com (Y.B.)}

\begin{document}

% Title Page with required iScience elements
\begin{titlepage}
\vspace*{2cm}

% Title
\noindent{\LARGE\textbf{EfficientMIL: Efficient Linear-Complexity MIL Method for WSI Classification}}

\vspace{1.5cm}

% Authors List
\noindent{\large
Chengying She\textsuperscript{1,2,5}, 
Chengwei Chen\textsuperscript{4,5}, 
Dongjie Fan\textsuperscript{3}, 
Lizhuang Liu\textsuperscript{1,6,*},
Chengwei Shao\textsuperscript{4,*},
Yun Bian\textsuperscript{4,*},
Ben Wang\textsuperscript{1,2}, 
Xinran Zhang\textsuperscript{1,2}
}

\vspace{0.8cm}

% Affiliations
\noindent{\small
\textsuperscript{1}Shanghai Advanced Research Institute, Chinese Academy of Sciences, Shanghai, China\\
\textsuperscript{2}University of Chinese Academy of Sciences, Beijing, China\\
\textsuperscript{3}North University of China, Taiyuan, China\\
\textsuperscript{4}Department of Radiology, Changhai Hospital, Shanghai, China\\
\textsuperscript{5}These authors contributed equally\\
\textsuperscript{6}Lead contact\\
\textsuperscript{*}Correspondence: liulz@sari.ac.cn (L.L.), chengweishaoch@163.com (C.S.), bianyun2012@foxmail.com (Y.B.)
}

\vspace{2cm}

% Graphical Abstract
\section*{GRAPHICAL ABSTRACT}
\begin{center}
\includegraphics[width=0.9\textwidth]{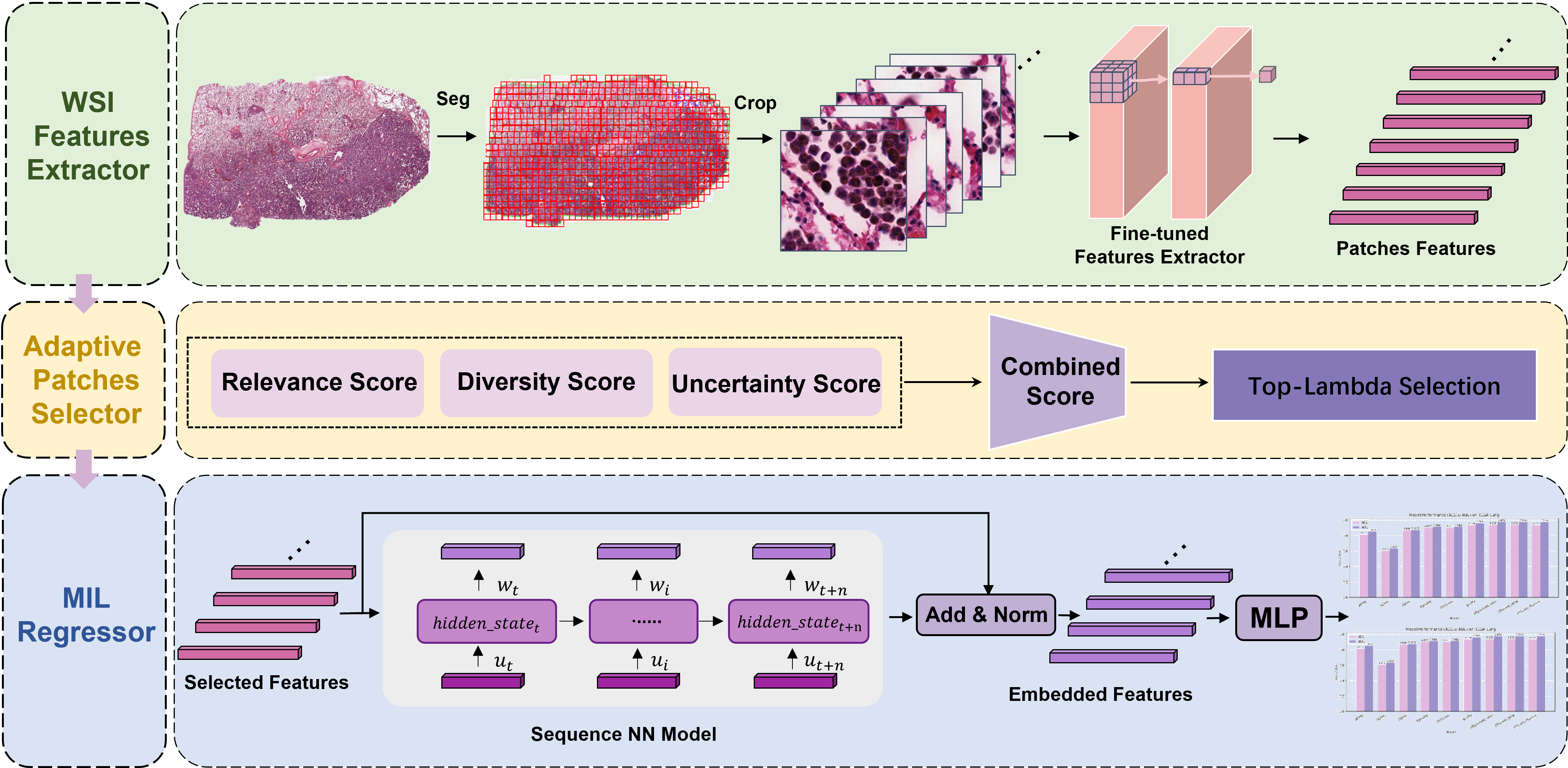}
\end{center}

\vspace{1.5cm}

% Highlights
\section*{HIGHLIGHTS}
\begin{itemize}
\item EfficientMIL introduces linear-complexity multiple instance learning for whole slide image classification, replacing quadratic-complexity attention mechanisms with efficient RNN-based sequence models
\item Adaptive patches selector (APS) intelligently identifies informative patches using relevance, diversity, and uncertainty criteria, significantly outperforming conventional selection strategies
\item EfficientMIL achieves state-of-the-art performance on TCGA-Lung and CAMELYON16 datasets while requiring substantially lower computational resources than attention-based methods
\end{itemize}

\vspace{1cm}

\end{titlepage}

\newpage

\maketitle

\section*{SUMMARY}

Whole slide images (WSIs) classification represents a fundamental challenge in computational pathology, where multiple instance learning (MIL) has emerged as the dominant paradigm. Current state-of-the-art (SOTA) MIL methods rely on attention mechanisms, achieving good performance but requiring substantial computational resources due to quadratic complexity when processing hundreds of thousands of patches. To address this computational bottleneck, we introduce EfficientMIL, a novel linear-complexity MIL approach for WSIs classification with the patches selection module Adaptive Patches Selector (APS) that we designed, replacing the quadratic-complexity self-attention mechanisms in Transformer-based MIL methods with efficient sequence models including RNN-based GRU, LSTM, and State Space Model (SSM) Mamba. EfficientMIL achieves significant computational efficiency improvements while outperforming other MIL methods across multiple histopathology datasets. On TCGA-Lung dataset, EfficientMIL-Mamba achieved AUC of 0.976 and accuracy of 0.933, while on CAMELYON16 dataset, EfficientMIL-GRU achieved AUC of 0.990 and accuracy of 0.975, surpassing previous state-of-the-art methods. Extensive experiments demonstrate that APS is also more effective for patches selection than conventional selection strategies.

\section*{KEYWORDS}

Computational Pathology, Multiple Instance Learning, Whole Slide Image, Linear Computational Complexity

\section*{INTRODUCTION}

Computational pathology has emerged as a transformative paradigm in precision oncology, fundamentally reshaping cancer diagnosis through automated analysis of gigapixel whole slide images (WSIs)\cite{vanderlaak2021_deep_learning_histopathology,song2023_artificial_intelligence_digital_computational_pathology,javed2020_cellular_community_detection,kather2016_multiclass_texture_analysis,ludwig2005_biomarkers_cancer}. These expansive images, frequently exceeding $100,000 \times 100,000$ pixels, capture intricate cellular structures in remarkable detail, offering unprecedented insights into disease pathology. Multiple deep learning-based studies have demonstrated that WSIs can yield information often imperceptible to human observers. However, the sheer scale and complexity of WSIs present formidable computational challenges that hinder seamless integration into routine clinical practice.

The standard workflow for WSI analysis involves segmenting tissue regions and cropping them into non-overlapping patches of fixed size, such as $256 \times 256$, at various magnifications\cite{li2021_dsmil,qu2022_dgmil,xu2024_prov_gigapath,jafarinia2024_snuffy}. Feature extraction is performed using pretrained convolutional neural network (CNN) or vision transformer (ViT) models \cite{campanella2019_clinical_grad,song2023_artificial_intelligence_digital_computational_pathology,qu2022_dgmil,dosovitskiy2021_vit,chen2022_scaling_vit}, converting each WSI into thousands of $d$-dimensional feature vectors, where $d$ is the dimension of each patch feature vector. Multiple instance learning (MIL) treats each slide as a "bag" containing multiple patch "instances", with only slide-level labels available during training\cite{maximilian2018_abmil,campanella2019_clinical_grad}.

Current state-of-the-art MIL methods for WSI analysis rely heavily on attention mechanisms to aggregate patch-level features into slide-level representations \cite{maximilian2018_abmil,shao2021_transmil,lu2021_clam,mao2025_camil}. While achieving high diagnostic accuracy, these approaches suffer from quadratic computational complexity with respect to the number of patches $N$, creating significant bottlenecks when processing large-scale WSI containing tens of thousands of patches. Recent efforts have explored alternatives such as sparse attention mechanisms. For example, Snuffy \cite{jafarinia2024_snuffy} reduces complexity to $O(\lambda N)$ by adopting sparse transformer patterns, based on efficient attention variants like Nyströmformer \cite{xiong2021_nystromformer}, where $\lambda \ll N$. However, computational demands remain high for resource-constrained clinical settings. Other approaches have investigated kernel attention transformers \cite{zheng2023_kat} and sparse transformer patterns \cite{child2019_sparse_transformer} to reduce computational complexity.

Additionally, current patches selection strategies often rely on simplistic approaches such as random sampling or selecting top-k patches based on single criteria, failing to capture the complexity of pathological analysis. These methods fail to account for the multifaceted factors that pathologists consider, including morphological diversity, diagnostic relevance, and prediction uncertainty across different regions. While some approaches like DGMIL \cite{qu2022_dgmil} and PAMIL \cite{liu2024_pamil} employ cluster-conditioned strategies and prototype learning, they still fall short of comprehensive strategies that integrate multiple factors for efficient informative patches selection. Recent work has explored attention-challenging approaches \cite{wang2024_attention_challenging} and diverse global representations \cite{zhu2024_dgr_mil} to improve patches selection and aggregation strategies.

To address these challenges, we propose EfficientMIL, a novel linear-complexity MIL architecture for WSI classification, replacing the quadratic-complexity self-attention mechanisms in Transformer-based MIL methods. We also propose a new patches selection module called adaptive patches selector (APS) that intelligently selects the most informative patches based on relevance, diversity, and uncertainty criteria. These designs achieve significant computational efficiency improvements while outperforming other MIL methods.

The purpose of this study is to (1) introduce a novel and efficient MIL framework to address the computational complexity issues inherent in traditional MIL methods and (2) propose a new patches selection module called adaptive patches selector (APS) that intelligently selects the most informative patches based on relevance, diversity, and uncertainty criteria and (3) explore the powerful applications of EfficientMIL on various WSI classification tasks.

\section*{RESULTS}

\subsection*{Overview of the EfficientMIL framework}

The EfficientMIL framework consists of three main components as illustrated in (Figure 1): (1) instance features extraction from WSI patches, (2) intelligent patches selection method named adaptive patches selector (APS), (3) efficient sequence models including RNN-based GRU \cite{dey2017_gru}, LSTM \cite{hochreiter1997_lstm}, and State Space Model (SSM) Mamba \cite{gu2023_mamba}. This design addresses the fundamental limitations of existing attention-based approaches by replacing quadratic complexity operations with linear-complexity sequential processing. To demonstrate the performance of the EfficientMIL framework on WSIs classification task, we evaluated it on two available datasets: TCGA-Lung \cite{cooper2018_tcga_pancancer} and CAMELYON16 \cite{bejnordi2017_camelyon16}, outperforming the SOTA methods. The results are shown in (Table 1) and (Table 2). Additionally, we evaluated it on several standard MIL datasets (MUSK1, MUSK2, ELEPHANT \cite{dietterich1997_mil,andrews2002_svm_mil}), results are shown in (Table 3). More details about the EfficientMIL framework can be found in the STAR Methods section.

\subsection*{EfficientMIL achieves superior performance on several WSI datasets}

We evaluated EfficientMIL on two available WSI datasets: TCGA-Lung and CAMELYON16 datasets \cite{cooper2018_tcga_pancancer,bejnordi2017_camelyon16}. Our approach consistently outperformed state-of-the-art methods while requiring significantly lower computational resources, as shown in (Table 1) and (Table 2). Moreover, (Figure 2) clearly illustrates the comparison between model performance and computational complexity.

On the TCGA-Lung dataset for lung cancer subtype classification (LUAD vs. LUSC), EfficientMIL-Mamba achieved the highest AUC of 0.976, while EfficientMIL-GRU achieved the highest accuracy of 0.938, representing improvements of 1.5\% and 0.4\% respectively over the previous best method Snuffy (Table 1). Notably, EfficientMIL-Mamba also demonstrated exceptional computational efficiency with only 3 million (M) FLOPs while maintaining competitive performance, significantly outperforming attention-based methods in computational efficiency.

Similar improvements were observed on the CAMELYON16 dataset for breast cancer metastasis detection (tumor vs. normal). EfficientMIL-GRU achieved the highest performance with AUC of 0.990 and accuracy of 0.975, surpassing the previous best method Snuffy by 3.4\% in AUC and 2.8\% in accuracy (Table 2).

\subsection*{Model visualization validates intelligent patches selection}

Qualitative analysis of patches selection results provides compelling evidence for the effectiveness of our APS module. On CAMELYON16 dataset samples, visualization of patch attention scores revealed high correspondence between selected high-scoring patches and tumor regions annotated by expert pathologists (Figure 4).

The heatmap visualization demonstrates that APS successfully identifies morphologically relevant regions while avoiding background areas and artifacts. High-scoring patches concentrate in areas with dense cellular structures and irregular morphology characteristic of metastatic regions.

Visualization of the scores from the intermediate-layer during inference revealed that the model achieved accurate classification and demonstrated robust patch-level tumor localization, with higher scores in tumoral regions (close to 1) and lower scores in non-tumoral areas (close to 0). This finding indicates that EfficientMIL can learn patch-level segmentation from weak supervision (slide-level labels) alone.

\subsection*{Strong performance on standard MIL benchmarks validates generalizability}

To demonstrate the broad applicability of our approach beyond WSI classification, we evaluated EfficientMIL on three standard MIL benchmark datasets: MUSK1, MUSK2, and ELEPHANT\cite{dietterich1997_mil,andrews2002_svm_mil}. Results show that EfficientMIL only performs slightly worse on MUSK1 compared to previous models, but outperforms previous algorithms on the other two datasets (Table 3). EfficientMIL-GRU achieved the highest AUC of 0.985 on ELEPHANT dataset, while EfficientMIL-LSTM and EfficientMIL-GRU both achieved perfect accuracy of 0.950 on MUSK2 dataset.

\subsection*{Adaptive patches selector outperforms conventional selection strategies}

To validate the effectiveness of our proposed adaptive patches selector (APS), we conducted comprehensive ablation studies comparing different patches selection strategies on the CAMELYON16 dataset using three EfficientMIL models with $\lambda=512$. The results are shown in (Figure 3A), which demonstrated that APS consistently outperformed simple selection strategies.

APS consistently outperformed simple selection strategies across different EfficientMIL models. Specifically, when EfficientMIL-GRU, APS achieved AUC of 0.980 compared to 0.955 for top-k selection and 0.944 of random-k selection, respectively representing 2.5\% and 3.6\% improvements. The accuracy improvement was also pronounced, for EfficientMIL-LSTM, APS achieved 0.988 compared to 0.940 for top-k selection and 0.966 for random-k selection, respectively representing 4.8\% and 1.6\% improvements.

\subsection*{Optimal parameters of APS enhance performance efficiency}

Investigation of different $\lambda$ values revealed important insights into the performance-efficiency trade-off (Figure 3B). Performance metrics gradually improved with larger $\lambda$ values, with ACC and AUC increasing slowly after $\lambda=512$. Beyond this point, performance gains diminished while computational costs (Training Time) continued to increase heavily.

\subsection*{Robustness across different WSI features extraction methods}

Evaluation across different WSI feature extraction methods demonstrated the robustness and generalizability of our approach in (Table 4). Using DSMIL \cite{li2021_dsmil} features (512-dimensional), EfficientMIL achieved mean AUC of 0.483 ± 0.019 and mean accuracy of 0.595 ± 0.000. With more advanced UNI2 \cite{chen2024_uni} features (1536-dimensional), performance significantly improved to mean AUC of 0.959 ± 0.030 and mean accuracy of 0.911 ± 0.076, demonstrating the importance of foundation models specifically trained for computational pathology.

\section*{DISCUSSION}

The EfficientMIL framework addresses critical limitations in current WSI classification methods by replacing computationally expensive attention mechanisms with efficient sequence model processing. Our results demonstrate that sequential modeling can effectively capture inter-patch dependencies while maintaining linear computational complexity, making the approach suitable for practical clinical deployment. The computational advantages of EfficientMIL are particularly significant for clinical applications. Traditional attention-based methods require substantial computational resources that may not be available in resource-constrained healthcare settings. EfficientMIL's linear complexity enables processing of arbitrarily large patch collections with constant memory per timestep, facilitating broader adoption of computational pathology tools.

The adaptive patches selector (APS) represents a significant advancement over simplistic selection strategies. The superior performance of APS stems from its multi-criteria optimization framework that simultaneously considers patch relevance, diversity, and uncertainty. By integrating these criteria, APS identifies truly informative patches that capture the complexity of pathological analysis, rather than relying on single-criterion rankings. The dynamic weight adjustment mechanism ensures balanced consideration of multiple factors throughout the selection process, enabling the framework to adapt to different pathological patterns and diagnostic requirements.

Our systematic evaluation of sequence architectures reveals that different variants offer distinct advantages. While LSTM and GRU provide robust bidirectional processing, Mamba offers exceptional computational efficiency as a state space model. Recent work has explored Mamba variants specifically for computational pathology \cite{yang2024_mambamil}, demonstrating the potential of state space models in this domain. This flexibility allows practitioners to select architectures based on specific computational constraints and performance requirements, making EfficientMIL adaptable to various clinical settings and resource limitations.

The linear-complexity design of EfficientMIL makes it particularly suitable for diverse MIL applications beyond computational pathology. The framework's ability to process large-scale instance collections efficiently opens possibilities for applications in drug discovery, where molecular compounds can be treated as bags of substructures, or in document classification, where documents serve as bags of words or sentences. The adaptive patches selector's multi-criteria optimization approach can be adapted to select informative instances in various domains, such as selecting relevant time points in time series analysis or identifying key regions in satellite imagery for environmental monitoring.

Furthermore, the model's demonstrated capability to learn fine-grained localization from weak supervision has significant implications for medical image analysis. This approach could be extended to other imaging modalities beyond histopathology, including radiology, dermatology, and ophthalmology, where precise localization of pathological regions is crucial for clinical decision-making. The combination of computational efficiency and localization capability positions EfficientMIL as a versatile framework for weakly supervised learning across multiple medical imaging domains.

Our results confirm that EfficientMIL can effectively leverage improvements in foundation models for computational pathology while maintaining its computational efficiency advantages. The consistent performance across different feature extraction methods highlights the framework's adaptability to evolving feature representation techniques. The linear-complexity design of EfficientMIL generalizes well beyond the computational pathology domain, making it suitable for more MIL applications, not limited to WSI classification, but also used for prognostic analysis.

\subsection*{Limitations of the study}

While EfficientMIL achieves strong accuracy with linear computational complexity, we acknowledge two practical limitations. First, because the core is a sequential model, GPU-level parallelism is limited relative to transformer-based methods that exploit highly parallel attention kernels. As a result, end-to-end training and inference times can be longer despite the lower algorithmic complexity. Second, the current design does not include explicit positional or spatial encoding for WSI patches, which means the model lacks direct awareness of inter-patch spatial relationships. This omission can weaken modeling of global tissue architecture and long-range spatial dependencies that are often informative in pathology.

\section*{STAR METHODS}

\subsection*{Key resources table}

\begin{table}[ht]
\centering
\scriptsize
\begin{tabular}{|l|l|l|}
\hline
\textbf{REAGENT or RESOURCE} & \textbf{SOURCE} & \textbf{IDENTIFIER} \\
\hline
\multicolumn{3}{|l|}{\textbf{Deposited data}} \\
\hline
The Cancer Genome Atlas (TCGA) & National Cancer Institute & \url{https://portal.gdc.cancer.gov/} \\
\hline
CAMELYON16 & Grand Challenge & \url{https://camelyon16.grand-challenge.org/} \\
\hline
MUSK1, MUSK2, ELEPHANT & Dietterich et al.\cite{dietterich1997_mil} & Standard MIL benchmarks \\
\hline
\multicolumn{3}{|l|}{\textbf{Software and algorithms}} \\
\hline
Python & Python Software Foundation & \url{https://www.python.org/} (v3.10.18) \\
\hline
PyTorch & Paszke et al.\cite{paszke2019_pytorch} & \url{https://pytorch.org/} (v2.1.1+cu118) \\
\hline
NumPy & Harris et al.\cite{harris2020_numpy} & \url{https://numpy.org/} \\
\hline
Scikit-learn & Pedregosa et al.\cite{pedregosa2011_scikit} & \url{https://scikit-learn.org/} (v1.6.1) \\
\hline
EfficientMIL & This paper & \url{https://github.com/chengyingshe/EfficientMIL} \\
\hline
\multicolumn{3}{|l|}{\textbf{Other}} \\
\hline
NVIDIA RTX 3090 GPU & NVIDIA Corporation & 2 $\times$ RTX 3090 (24GB VRAM each) \\
\hline
Ubuntu & Canonical Ltd. & Ubuntu 20.04.6 LTS \\
\hline
CUDA & NVIDIA Corporation & CUDA 11.8 \\
\hline
\end{tabular}
\end{table}

\subsection*{Experimental model and study participant details}

This study utilized publicly available datasets without direct human participant involvement. TCGA-Lung dataset contains 1,042 WSIs from patients with lung adenocarcinoma (LUAD, n=530) and lung squamous cell carcinoma (LUSC, n=512). CAMELYON16 dataset includes 399 WSIs from breast cancer patients with normal tissue (n=240) and tumor tissue (n=159). All data were previously collected under appropriate institutional review board approvals and informed consent procedures as described in the original publications.

\subsection*{Method details}

\subsubsection*{WSI preprocessing and feature extraction}

For each WSI, tissue regions were segmented using Otsu thresholding \cite{otsu1979_otsu} and cropped into non-overlapping fixed size patches at 20$\times$ magnification (i.e, $256 \times 256$). Feature extraction was performed using the foundation model UNI2 \cite{chen2024_uni} yielding 1536-dimensional vectors. Recent advances in foundation models for computational pathology\cite{elnahhas2025_stamp} have demonstrated the effectiveness of such feature extractors for downstream tasks. Patches with insufficient tissue content were filtered using adaptive thresholding.

\subsubsection*{Adaptive patches selector implementation}

The APS module combines three criteria with fixed weights and performs single-pass selection by scoring all candidate patches, then selecting the top-$\lambda$ patches. Given a bag containing $N$ patches with features $\{f_1, f_2, \ldots, f_N\}$ where each $f_i \in \mathbb{R}^d$ represents the $d$-dimensional feature vector of patch $i$, and instance logits $\mathbf{c}_i \in \mathbb{R}^C$ from the instance classifier for patch $i$, we define three scoring criteria:

\noindent\textbf{Relevance Score.} The relevance score measures the diagnostic importance of each patch based on its classification confidence. We convert instance logits to probabilities and take the maximum class probability as the relevance score:
\begin{equation}
S_{rel}(f_i) = \max_{c \in \{1,\dots,C\}} p_c(f_i), \quad p(f_i) =
\begin{cases}
\sigma(\mathbf{c}_i) & C=1 \\
\mathrm{softmax}(\mathbf{c}_i) & C>1
\end{cases}
\end{equation}
where $C$ is the number of classes, $\sigma(\cdot)$ is the sigmoid function for binary classification, $\mathrm{softmax}(\cdot)$ is the softmax function for multi-class classification, and $p_c(f_i)$ represents the predicted probability of patch $i$ belonging to class $c$.

\noindent\textbf{Diversity Score.} The diversity score encourages selection of morphologically dissimilar patches to ensure comprehensive tissue representation. We compute cosine similarity between all patch pairs and define diversity as the complement of average similarity:
\begin{equation}
S_{div}(f_i) = 1 - \frac{1}{N-1}\sum_{j\neq i} S_{ij}
\end{equation}
where $\mathbf{x}_k = \mathbf{f}_k/\lVert \mathbf{f}_k\rVert_2$ is the L2-normalized feature vector of patch $k$, $\mathbf{S} = \mathbf{X}\mathbf{X}^\top$ is the $N \times N$ cosine similarity matrix with elements $S_{ij} = \mathbf{x}_i^T \mathbf{x}_j$, and $N$ is the total number of patches in the bag.

\noindent\textbf{Uncertainty Score.} The uncertainty score captures prediction entropy to identify challenging regions that may benefit from additional attention:
\begin{equation}
S_{unc}(f_i) = -\sum_{c=1}^{C} p_c(f_i) \log\big(p_c(f_i)+10^{-8}\big)
\end{equation}
where $p_c(f_i)$ is the predicted probability of patch $i$ belonging to class $c$, and the small constant $10^{-8}$ prevents numerical instability when $p_c(f_i) = 0$.

\noindent\textbf{Final Score and Selection.} The final selection score combines all three criteria with fixed weights:
\begin{equation}
S_{final}(f_i) = w_{rel}\, S_{rel}(f_i) + w_{div}\, S_{div}(f_i) + w_{unc}\, S_{unc}(f_i)
\end{equation}
where $w_{rel}=1.0$, $w_{div}=0.3$, and $w_{unc}=0.3$ are the fixed weights for relevance, diversity, and uncertainty respectively. We select the top-$\lambda$ patches ranked by $S_{final}(f_i)$ and compute attention-like weights as $\mathrm{softmax}(S_{final})$ over all patches for downstream processing. The parameter $\lambda$ (denoted as \texttt{big\_lambda} in the implementation) is configurable; we use $\lambda=512$ for WSI experiments unless otherwise specified which is the optimal setting in the experiments results (Table 3). The computational complexity is $\mathcal{O}(N^2 d)$ for diversity computation due to the cosine similarity matrix, with additional $\mathcal{O}(N\log N)$ for sorting and selection.

\subsubsection*{Sequence model architecture configurations}

We evaluate three efficient sequence encoders, each followed by residual connection, layer normalization, global average pooling, and a linear classifier. The processing pipeline for each sequence model variant is:

\begin{itemize}
\item \textbf{Bidirectional LSTM}: hidden size 768, 2 layers, dropout 0.1
\item \textbf{Bidirectional GRU}: hidden size 768, 2 layers, dropout 0.1  
\item \textbf{SSM Mamba}: depth 8 blocks, state dimension 32, convolution kernel 4, expansion 2, dropout 0.1
\end{itemize}

After selecting $\lambda$ patch features via APS, the sequence processing pipeline operates as follows:

\textbf{Step 1: Sequence Encoding.} The selected patch features $\mathbf{X} \in \mathbb{R}^{1 \times \lambda \times d}$ are processed through the chosen sequence encoder to capture inter-patch dependencies:
\begin{equation}
\mathbf{H} = \text{Encoder}(\mathbf{X})
\end{equation}
where $\mathbf{H} \in \mathbb{R}^{1 \times \lambda \times d}$ represents the encoded features. For bidirectional RNNs (LSTM/GRU), the encoder processes sequences in both forward and backward directions, while Mamba uses state space modeling for efficient long-range dependency capture.

\textbf{Step 2: Residual Connection and Normalization.} The encoded features are combined with the original input through a residual connection, followed by LayerNorm for LSTM and GRU and RMSNorm for Mamba:
\begin{equation}
\mathbf{H}_{res} = \text{Norm}(\mathbf{H} + \mathbf{X})
\end{equation}
where the residual connection helps preserve gradient flow and the layer normalization stabilizes training.

\textbf{Step 3: Global Pooling and Classification.} The final bag representation is obtained through global average pooling:
\begin{equation}
\mathbf{z}_{bag} = \frac{1}{\lambda} \sum_{i=1}^{\lambda} \mathbf{H}_{res}[i]
\end{equation}
where $\mathbf{z}_{bag} \in \mathbb{R}^{1 \times d}$ is the bag-level representation. The final classification logits are computed as:
\begin{equation}
\hat{\mathbf{y}} = \mathbf{W}_{cls} \mathbf{z}_{bag} + \mathbf{b}_{cls}
\end{equation}
where $\mathbf{W}_{cls} \in \mathbb{R}^{C \times d}$ and $\mathbf{b}_{cls} \in \mathbb{R}^{C}$ are the classification layer parameters, and $C$ is the number of classes.

Our training objective uses binary cross-entropy with logits (BCEWithLogits) on two terms: the bag-level prediction and the max-pooled instance prediction, combined with equal weights, plus L2 regularization on parameters. $\lambda_{L2}$ is set to $10^{-4}$:
\begin{equation}
  \begin{cases}
    L_{main} = \tfrac{1}{2}\, \mathrm{BCELogits}(\hat{y}_{bag}, y) + \tfrac{1}{2}\, \mathrm{BCELogits}(\max_i\, \hat{y}_{inst,i}, y), \\ 
    L_{total} = L_{main} + \lambda_{L2}\, \lVert\theta\rVert_2^2
  \end{cases}
\end{equation}

\subsubsection*{Training procedure} 

Models were trained using Adam optimizer \cite{loshchilov2019_adamw} with learning rate $2\times10^{-4}$ (betas $(0.5, 0.9)$) and weight decay $1\times10^{-5}$. We used a cosine annealing scheduler with minimum learning rate $5\times10^{-6}$. Unless stated otherwise, we trained for 50 epochs with early stopping (patience 5 epochs without validation improvement). Batch size was set to 1. For fair comparison across different models, we used a consistent train/validation split ratio of 4:1 (80\%/20\%) for all experiments. The random seed was fixed to 42 for reproducibility.

\subsection*{Quantification and statistical analysis}

Performance evaluation used a consistent train/validation split ratio of 4:1 (80\%/20\%) across all models to ensure fair comparison. The same dataset partitioning and training parameters were applied to all baseline methods and our proposed EfficientMIL variants. Area under the ROC curve (AUC) and accuracy (ACC) were used as primary evaluation metrics for binary classification tasks.

Computational efficiency was measured using FLOPs (floating-point operations), model size (in megabytes), and memory usage (in megabytes) on NVIDIA RTX 3090 GPUs. Inference times were averaged over 100 runs after warm-up periods.

\section*{RESOURCE AVAILABILITY}

\subsection*{Lead contact}

Requests for further information and resources should be directed to and will be fulfilled by the lead contact, Lizhuang Liu (liulz@sari.ac.cn).

\subsection*{Materials availability}

This study did not generate new physical materials. All computational models and algorithms are described in sufficient detail for reproduction.

\subsection*{Data and code availability}

\begin{itemize}
    \item The TCGA-Lung dataset is available through The Cancer Genome Atlas portal (\url{https://portal.gdc.cancer.gov/}). The CAMELYON16 dataset is available at \url{https://camelyon16.grand-challenge.org/}. MIL benchmark datasets are available at \url{https://www.uco.es/grupos/kdis/momil/}. All datasets are publicly accessible as of the date of publication.
    \item All original code for EfficientMIL implementation, including the adaptive patches selector and sequence model architectures, has been deposited at GitHub under repository \url{https://github.com/chengyingshe/EfficientMIL} and will be made publicly available upon publication acceptance.
    \item Any additional information required to reanalyze the data reported in this paper is available from the lead contact upon request.
\end{itemize}

\section*{ACKNOWLEDGMENTS}

The authors thank the TCGA and CAMELYON16 consortiums for providing publicly available datasets. We acknowledge computational resources provided by the Industrial Artificial Intelligence Team of Shanghai Advanced Research Institute.

\section*{AUTHOR CONTRIBUTIONS}

L.L., C.S., and Y.B. conceived the project and provided supervision and funding acquisition; C.S. and C.C. developed the EfficientMIL framework; C.S., B.W. and X.Z. collected and preprocessed the WSI datasets, performed the experiments, analyzed the results and wrote the manuscript. All authors read and approved the final manuscript.

\section*{DECLARATION OF INTERESTS}

The authors declare no competing interests.

\section*{MAIN FIGURE TITLES AND LEGENDS}

\noindent\includegraphics[width=\linewidth]{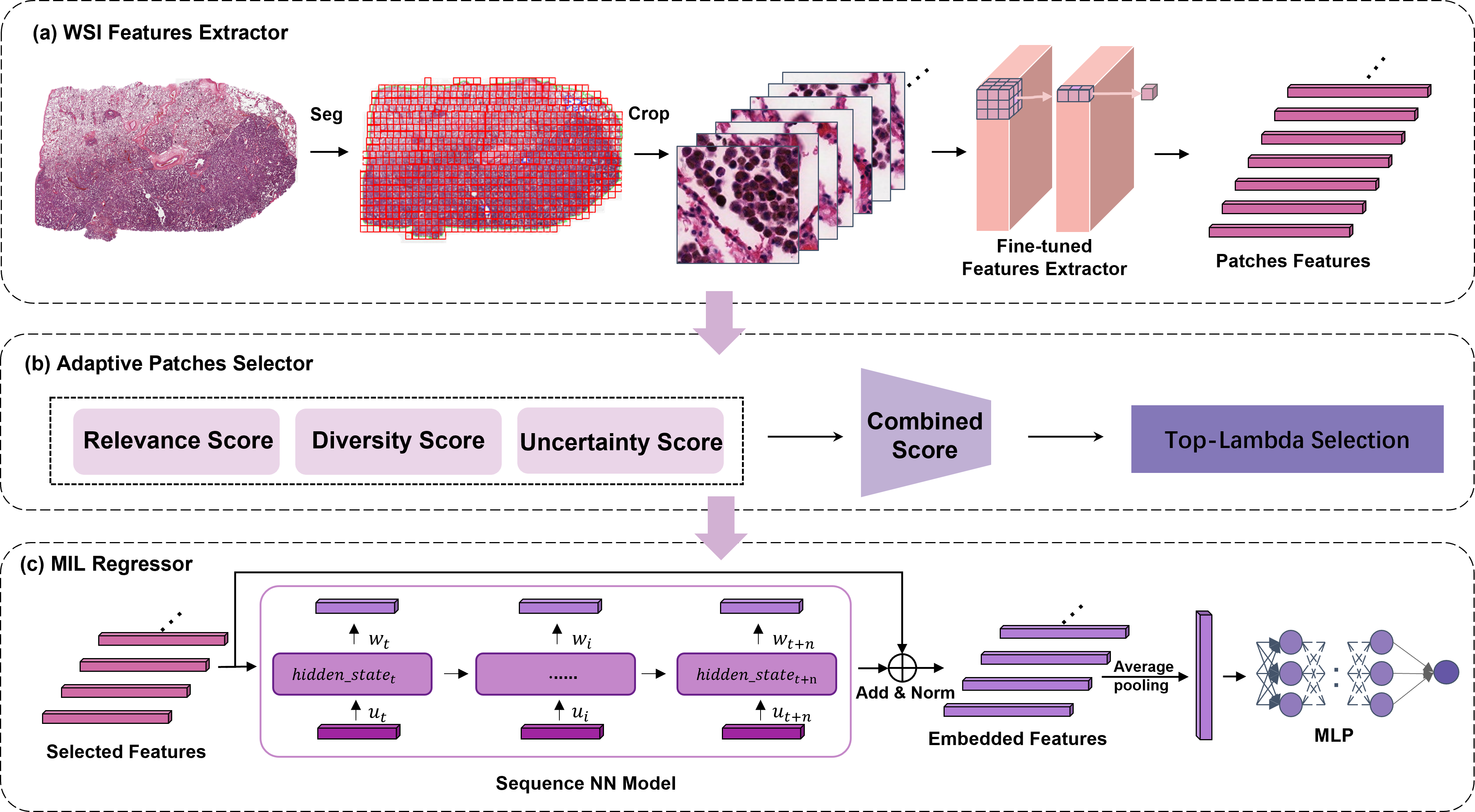}

\subsection*{Figure 1. An overview of the EfficientMIL pipeline}

\noindent\includegraphics[width=\linewidth]{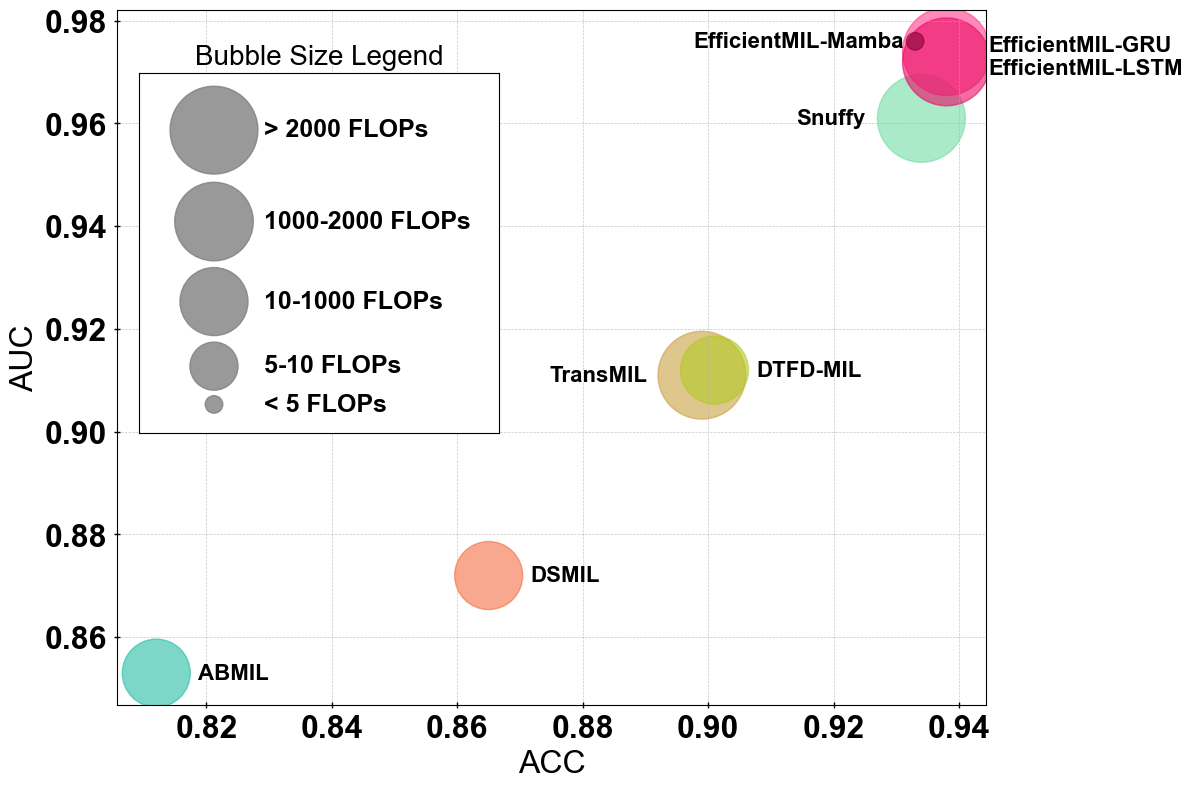}

Schematic overview of the EfficientMIL framework showing the three main components: (1) WSI feature extractor to extract instance features from WSIs, (2) Adaptive patches selector to select the most informative patches, (3) MIL regressor with sequence neural network modules (LSTM, GRU, Mamba) to model the inter-patch dependencies and predict the final label.

\subsection*{Figure 2. Model Performance (ACC and AUC) vs. computational complexity (FLOPs) trade off}

Performance comparison of different MIL methods on CAMELYON16 dataset with $\lambda=512$ for EfficientMIL models.

\noindent\includegraphics[width=\linewidth]{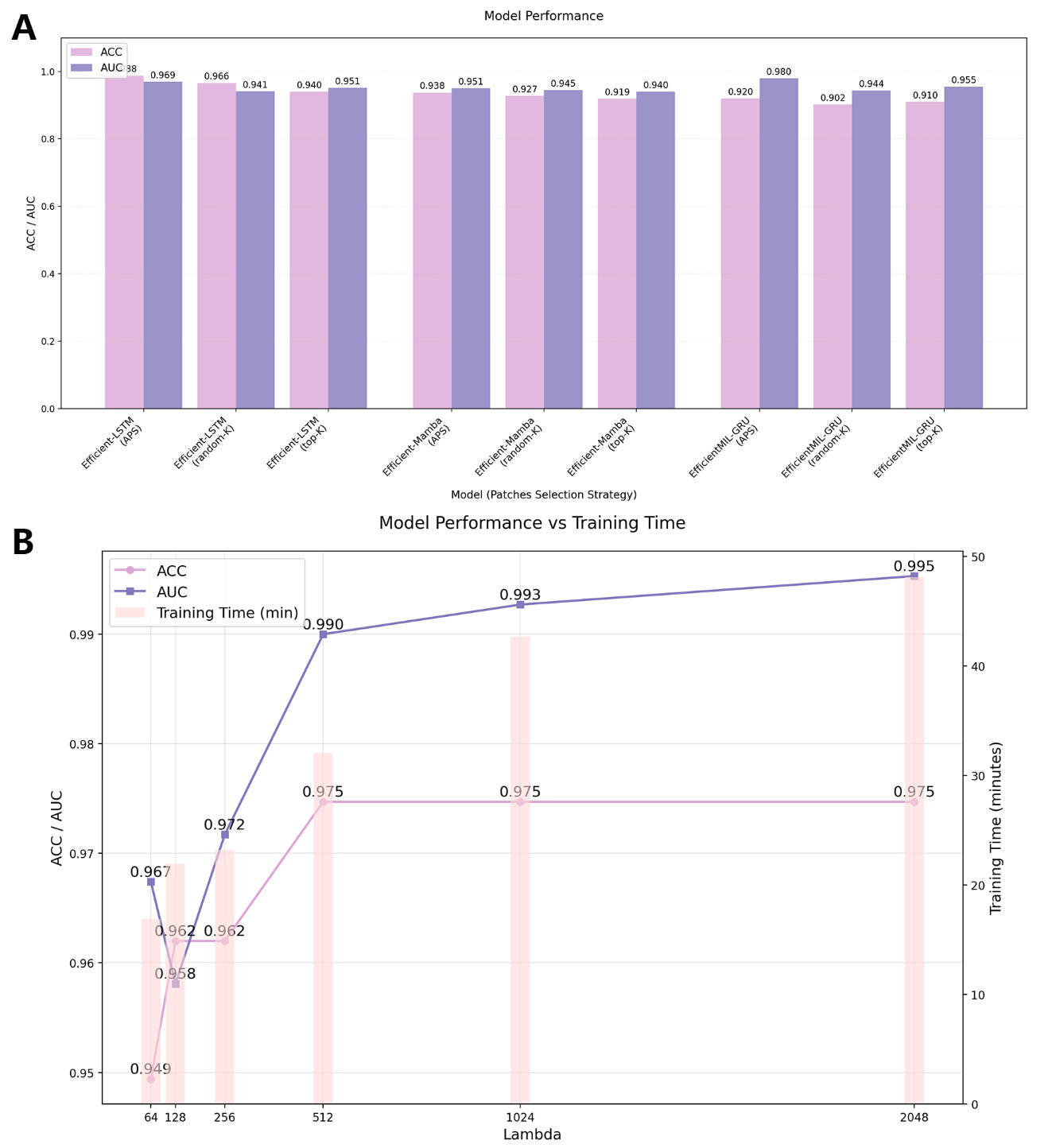}

\subsection*{Figure 3. Model Performance with different patches selection strategies and different number of selected patches}

(A) Comparison of model performance on CAMELYON16 dataset using all three EfficientMIL models with different patches selection strategies with $\lambda=512$, and (B) comparison of model performance and computational resources consumption (Training Time) on CAMELYON16 dataset with different numbers of selected patches ($\lambda=64,128,256,512,1024,2048$).

\noindent\includegraphics[width=\linewidth]{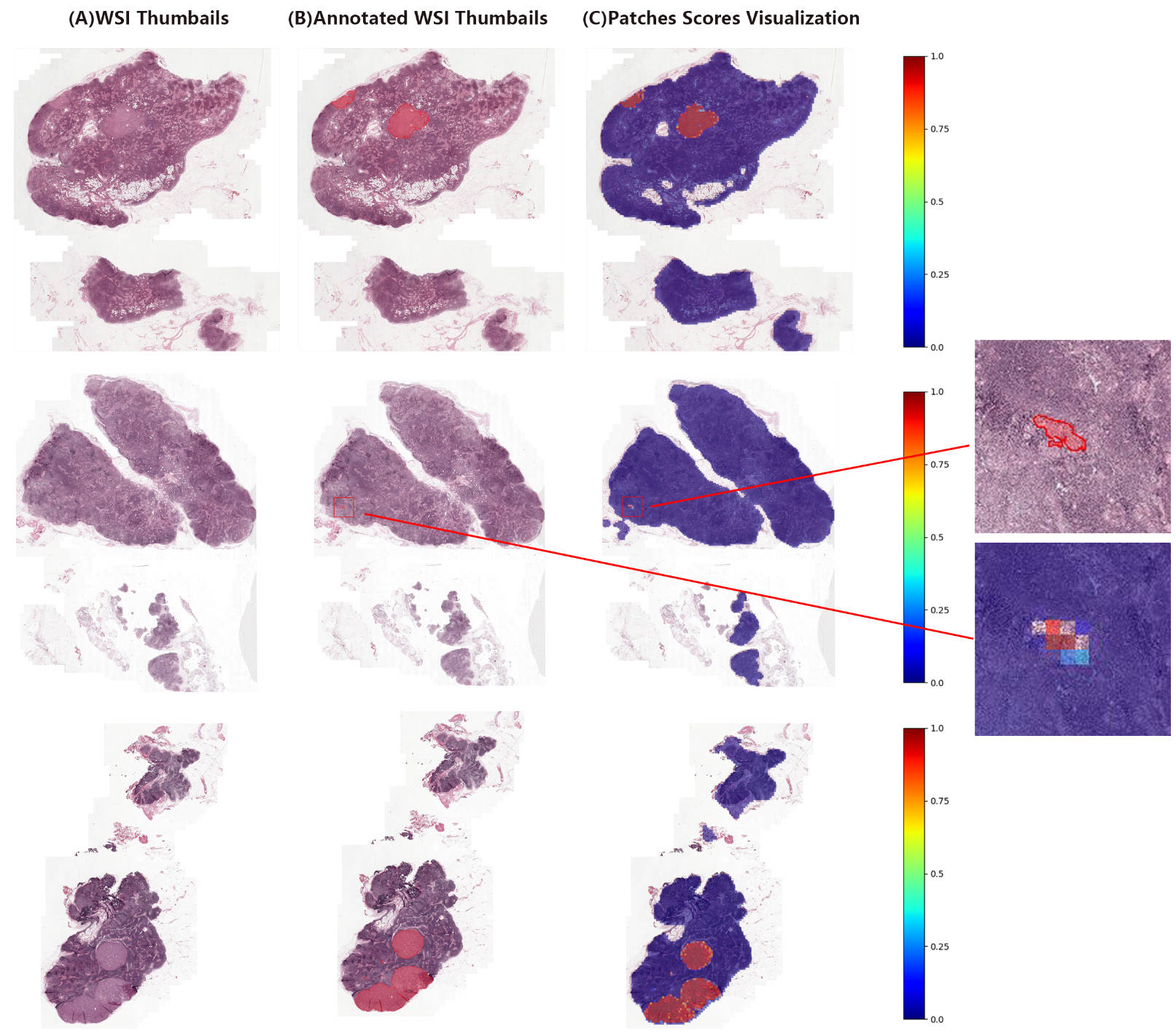}

\subsection*{Figure 4. Visualization of adaptive patches selection results}

Qualitative analysis of patches scores on CAMELYON16 dataset samples. (A) Original WSI thumbnails. (B) Ground truth tumor annotation with red regions indicating tumor regions. (C) Patches scores heatmap generated by EfficientMIL-GRU showing high correspondence between high-scoring patches and tumor regions.

\section*{MAIN TABLES, INCLUDING TITLES AND LEGENDS}

\subsection*{Table 1. Results on TCGA-Lung dataset (LUAD vs. LUSC)}

\begin{table}[ht]
\centering
\begin{tabular}{|l|l|c|c|c|c|}
\hline
\textbf{Method} & \textbf{Reference} & \textbf{AUC} & \textbf{ACC} & \textbf{FLOPs (M)} & \textbf{Model Size (MB)}\\
\hline
ABMIL & Ilse et al. \cite{maximilian2018_abmil} & 0.853 & 0.812 & 396 & \textbf{1.52} \\
DGMIL & Qu et al. \cite{qu2022_dgmil} & 0.632 & 0.601 & 2369 & 9.05 \\
DSMIL & Li et al. \cite{li2021_dsmil} & 0.872 & 0.865 & 216 & 0.86 \\
TransMIL & Shao et al. \cite{shao2021_transmil} & 0.911 & 0.899 & 2989 & 11.20 \\
DTFD-MIL & Zhang et al. \cite{zhang2022_dtfd_mil} & 0.912 & 0.901 & 921 & 4.02 \\
Snuffy & Jafarinia et al. \cite{jafarinia2024_snuffy} & 0.961 & 0.934 & 90803 & 360.40 \\
\hline
EfficientMIL-LSTM & This work & 0.974 & \textbf{0.938} & 14527 & 108.13 \\
EfficientMIL-GRU & This work & 0.972 & \textbf{0.938} & 10898 & 81.11 \\
EfficientMIL-Mamba & This work & \textbf{0.976} & 0.933 & \textbf{3} & 487.39 \\
\hline
\end{tabular}
\end{table}

Performance comparison on TCGA-Lung dataset (LUAD vs. LUSC) using UNI2 \cite{chen2024_uni} as WSI features extractor with $\lambda=512$. FLOPs in megaflops (M), model size in megabytes (MB).

\subsection*{Table 2. Results on CAMELYON16 dataset (normal vs. tumor)}

\begin{table}[ht]
\centering
\begin{tabular}{|l|c|c|c|}
\hline
\textbf{Method} & \textbf{AUC} & \textbf{ACC} & \textbf{Inference Time (ms)} \\
\hline
ABMIL & 0.873 & 0.835 & \textbf{0.24} \\
DGMIL & 0.659 & 0.620 & 0.37 \\
DSMIL & 0.889 & 0.875 & 0.49 \\
TransMIL & 0.926 & 0.906 & 1.46 \\
DTFD-MIL & 0.925 & 0.912 & 2.10 \\
Snuffy & 0.956 & 0.947 & 16.60 \\
\hline
EfficientMIL-LSTM & 0.953 & 0.937 & 28.74 \\
EfficientMIL-GRU & \textbf{0.990} & \textbf{0.975} & 20.99 \\
EfficientMIL-Mamba & 0.933 & 0.823 & 12.42 \\
\hline
\end{tabular}
\end{table}

Results on CAMELYON16 dataset (normal vs. tumor) using UNI2 \cite{chen2024_uni} as WSI features extractor with $\lambda=512$. Inference time is measured while the models inference on a single WSI (i.e., $batch\_size=1$).

\subsection*{Table 3. Results on standard MIL datasets}

\begin{table}[ht]
\centering
\begin{tabular}{|l|c|c|c|c|c|c|}
\hline
\textbf{Method} & \multicolumn{2}{c|}{\textbf{MUSK1}} & \multicolumn{2}{c|}{\textbf{MUSK2}} & \multicolumn{2}{c|}{\textbf{ELEPHANT}} \\
\hline
 & \textbf{AUC} & \textbf{ACC} & \textbf{AUC} & \textbf{ACC} & \textbf{AUC} & \textbf{ACC} \\
\hline
ABMIL & 0.838 & 0.833 & 0.917 & 0.900 & 0.923 & 0.925 \\
DSMIL & \textbf{0.974} & 0.944 & 0.901 & 0.900 & 0.905 & 0.875 \\
Snuffy & 0.950 & 0.944 & 0.989 & 0.950 & 0.980 & 0.925 \\
\hline
EfficientMIL-LSTM & 0.938 & 0.944 & 0.945 & 0.950 & 0.983 & \textbf{0.950} \\
EfficientMIL-GRU & 0.938 & 0.889 & \textbf{0.960} & \textbf{0.951} & \textbf{0.985} & \textbf{0.950} \\
EfficientMIL-Mamba & 0.948 & 0.944 & 0.857 & 0.850 & 0.920 & 0.925 \\
\hline
\end{tabular}
\end{table}

Performance comparison on standard MIL benchmark datasets. EfficientMIL variants demonstrate competitive performance across diverse domains, confirming the generalizability of our linear-complexity approach beyond computational pathology.

\subsection*{Table 4. Comparison of different WSI feature extractors}

\begin{table}[ht]
\centering
\begin{tabular}{|l|c|c|c|c|c|}
\hline
\textbf{Feature Extractor} & \textbf{Patch Size} & \textbf{Feature Dim} & \textbf{ACC} & \textbf{AUC} \\
\hline
ResNet50\cite{he2016_resnet} & 256 & 1024 & 0.603 ± 0.015 & 0.473 ± 0.048 \\
DSMIL\cite{li2021_dsmil} & 224 & 512 & 0.595 ± 0.000 & 0.483 ± 0.019 \\
GPFM\cite{ma2025_gpfm} & 256 & 1024 & 0.907 ± 0.053 & 0.942 ± 0.035 \\
UNI2\cite{chen2024_uni} & 256 & 1536 & \textbf{0.911 ± 0.076} & \textbf{0.959 ± 0.030} \\
\hline
\end{tabular}
\end{table}

Comparison of EfficientMIL performance (mean ± standard deviation across EfficientMIL-GRU, EfficientMIL-LSTM, and EfficientMIL-Mamba) using different WSI feature extractors on CAMELYON16 dataset with $\lambda=512$. ResNet50 is pretrained on ImageNet, DSMIL is the WSI features extraction method using self-supervised contrastive learning on patches extracted at multiple magnifications followed by a pyramidal concatenation strategy from the paper \cite{li2021_dsmil}. GPFM and UNI2 are both pathological foundation models. Results demonstrate robustness across feature extraction methods while showing improved performance with more advanced foundation models.

\bibliography{references}

\end{document}